\documentclass[dvipsnames]{article} 
\usepackage{iclr2023_conference,times}

\usepackage{amsmath,amsfonts,bm}

\def\eqref#1{equation~\ref{#1}}

\def\1{\bm{1}}

\def\mP{{\bm{P}}}

\DeclareMathAlphabet{\mathsfit}{\encodingdefault}{\sfdefault}{m}{sl}
\SetMathAlphabet{\mathsfit}{bold}{\encodingdefault}{\sfdefault}{bx}{n}

\definecolor{goor}{HTML}{DB4437}
\definecolor{gooy}{HTML}{F4B400}
\definecolor{goob}{HTML}{4285F4}
\definecolor{goog}{HTML}{0F9D58}

\usepackage{hyperref}
\usepackage{url}
\usepackage{nicefrac}
\usepackage{booktabs}
\usepackage{amsfonts}
\usepackage{bold-extra}
\usepackage{multirow}
\usepackage{multicol}
\usepackage{xspace}
\usepackage{xcolor}
\usepackage{makecell}
\usepackage{subfigure}
\usepackage{graphicx}

\usepackage{url}
\usepackage{amsmath}
\usepackage{tabularx}
\usepackage{dcolumn,caption}
\usepackage{arydshln}
\usepackage{tikz}
\usepackage{pgfplots}
\usepgfplotslibrary{statistics}
\usetikzlibrary{plotmarks}

\usepackage{algorithm,algorithmicx,algpseudocode}
\usepackage{wrapfig}

\usepackage{pifont}
\usepackage{bm}
\usepackage{bbm}
\usepackage{enumitem}

\usepackage{arydshln}
\usepackage{amssymb}
\usepackage{amsbsy}
\usepackage{tcolorbox}
\usepackage{adjustbox}
\usepackage{pifont}

\usepackage[title]{appendix}

\newcommand\ours{\textsc{SoLiS}\xspace}
\newcommand\poet{\textsc{PoEt}}
\newcommand\tapex{\textsc{TaPEx}}
\newcommand\ourtask{MathExp\xspace}

\definecolor{sqlpurple}{RGB}{115,50,164}

\newcommand{\sqlcorrect}[1]{{\color{ForestGreen}{\cmark}}}
\newcommand{\sqlwrong}[1]{{\color{red}{\xmark}}}

\newcommand{\cmark}{\ding{51}}
\newcommand{\xmark}{\ding{55}}

\title{Reflection of Thought: Inversely Eliciting Numerical Reasoning in Language Models via Solving Linear Systems}

\author{
Fan Zhou\textsuperscript{\rm 1}$\thanks{Equal contributions.}$~,
Haoyu Dong\textsuperscript{\rm 2}$^{*}$$\thanks{Corresponding author.}$~,
Qian Liu\textsuperscript{\rm 3},
Zhoujun Cheng\textsuperscript{\rm 1},
Shi Han\textsuperscript{\rm 2},
Dongmei Zhang\textsuperscript{\rm 2}\\
\textsuperscript{\rm 1}{Shanghai Jiao Tong University,}
\textsuperscript{\rm 2}{Microsoft Research Asia,}
\textsuperscript{\rm 3}{Sea AI Lab}\\
\{zhoufan98,blankcheng\}@sjtu.edu.cn,
\{hadong, shihan, dongmeiz\}@microsoft.com,
liuqian@sea.com}

\begin{document}

\maketitle

\begin{abstract}

Numerical reasoning over natural language has been a long-standing goal for the research community.
However, cutting-edge language models have proven difficult to reliably generalize to a broad range of numbers, although they have shown proficiency in reasoning over common and simple numbers.
In this paper, we propose a novel method to elicit and exploit the numerical reasoning knowledge hidden in pre-trained language models using simple anchor numbers.
Concretely, we first leverage simple numbers as anchors to probe the implicitly inferred arithmetic expressions from language models, and then explicitly apply the expressions on complex numbers to get corresponding answers.
To inversely elicit arithmetic expressions, we transform and formulate the task as an analytically solvable linear system.
Experimental results on several numerical reasoning benchmarks demonstrate that our approach significantly improves numerical reasoning capabilities of existing LMs.
More importantly, our approach is training-free and simply works in the inference phase, making it highly portable and achieving consistent performance benefits across a variety of language models (GPT-3, T5, BART, etc) in all zero-shot, few-shot, and fine-tuning scenarios.

\end{abstract}

\section{Introduction}\label{sec:intro}
Language Models (LMs) have demonstrated great success on a wide range of natural language tasks~\citep{devlin2018bert,brown2020language,chowdhery2022palm}, and recent works even explore to use LMs as a general-purpose interface for diverse modalities~\citep{hao2022language,xie2022unifiedskg,he2022masked}. But when it comes to reasoning about numbers, the crucial parts of text, tables, and knowledge bases, the performance of LMs slumps.
Even rational numbers, a small subset of real numbers, readily constitute an infinite space that cannot be completely covered by pre-training corpora, hence posing a significant obstacle to LMs.
Recent works have shown strong context understanding capabilities of LMs in numerical reasoning datasets~\citep{dua2019drop,cobbe2021GSM8K}, but LMs are still far from being robust on end-to-end numerical calculation:
as numbers get larger and more complicated, the likelihood of failure for LMs increases, e.g., it is difficult for them to calculate the result of $8,534.5+17.85$;
even for number additions between small numbers, e.g., $512+128$ and $513+129$, LMs are not stable enough to consistently produce the correct result.
Similar observations are also reported by~\citet{razeghi2022impact}, showing that LMs struggle to conduct end-to-end calculations on numbers that rarely appear in pre-training corpora.

\begin{figure}[t]
    \centering
    \includegraphics[width=0.98\textwidth]{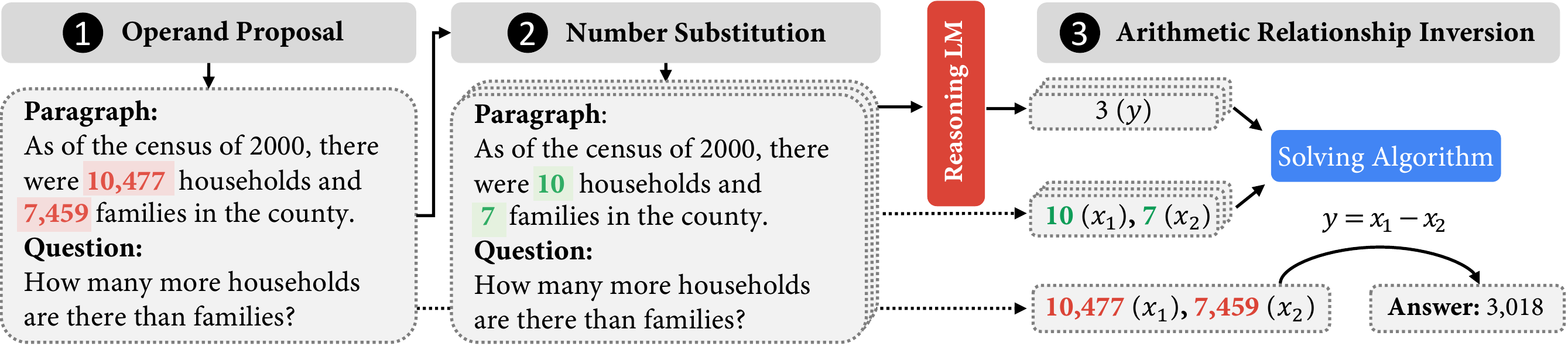}
    \caption{The illustration of our proposed framework, which elicits numerical reasoning in language models via \textbf{So}lving \textbf{Li}near \textbf{S}ystems (\ours).}
    \label{fig:model_intro}
\end{figure}

Fortunately, by reverse thinking, we have a positive point of view: with the exact same context, LMs are significantly more accurate and stable on simple numbers - typically small integers that appear frequently in the pre-training corpora - than complex numbers, indicating that LMs have a strong aptitude for bearing arithmetic results of simple numbers in mind during pre-training.
This motivates us to \textit{leverage simple numbers as ``anchors'' to probe the implicitly inferred arithmetic expressions from language models and then explicitly apply the expressions on complex numbers}.
Specifically, as Figure~\ref{fig:model_intro} illustrates, when detecting complex numbers (e.g., $10,477$, $7,459$) that are challenging for LMs, to first replace them by anchor numbers (e.g., $10$, $7$) and use LMs to output answers (e.g., $3$) that are more much accurate than complex numbers, then inversely elicit the hidden arithmetic relationship (e.g., $x_1-x_2$) implicitly inferred by LMs through these anchor inputs and outputs, and finally explicitly applying the arithmetic relationship on the original complex numbers ($10,477-7,459$) to produce the answer ($3,018$). In this way, our method combines the advances of LMs on understanding complex context and memorizing simple numbers towards reliable numerical reasoning. 

This paper introduces reflection of thought, a new idea of eliciting the numerical reasoning knowledge hidden in pre-trained LMs through probing with simple anchor numbers.
Reflection of thought, in principle, allows models to unveil the underlying reasoning process by varying inputs at test time, so it does not need any extra training or labeled data.
To inversely elicit arithmetic relationships in LMs through anchor numbers, we propose \textsc{SoLiS}, a novel method to transform and formulate this problem to a linear system that can be directly solved in an analytic way. 
To promote robustness to mistakes introduced by LMs, search-based and heuristic-based methods are further devised.
Experimental results on several representative numerical reasoning datasets demonstrate that \textsc{SoLiS} significantly advances cutting-edge LMs.
Importantly, our framework simply works in the inference phase without extra training or labeled data, making it highly portable to different kinds of LMs and achieving consistent gains over various LMs in zero-shot, few-shot and fine-tuning scenarios.

\section{Preliminary Study}\label{sec:preliminary}

In this section, we will first demonstrate the brittleness of language models' ability on arithmetically-related tasks. 
Unlike arithmetic benchmarks such as AddSub or MultiArith~\citep{roy2015solving} which contain natural language context for each sample, we directly generate and feed the arithmetic expressions and test the performance on language models. 
This is done to reduce potential perturbing factors and highlight the models' calculating ability.
We impose constraints on the complexity of the expressions: we only study the four fundamental operations, and demand no more than $4$ operands, where each operand's integer range is less then $10,000$ and floating point precision is less than $4$. 
To conduct a systematic investigation, we first produce $\mathbb{F}$ which represents the set of all the expressions satisfying our constraints.
We randomly sample numbers within the limits of range and precision as the operands.
For one expression $f \in \mathbb{F}$ with a specified range and precision, we randomly generate $50$ samples. We evaluate the language model on these samples and denote this synthesized task as \ourtask which stands for \textbf{Math} \textbf{Exp}ressions.

\begin{wrapfigure}[12]{R}{0.5\textwidth}
\centering
\begin{tikzpicture}
\small{
\begin{axis}[
at={(0,0)},
width=.27\textwidth, height=.22\textwidth ,
xtick={1,2,3,4},
xticklabels={$0$, $1$, $2$, $3$},
ytick={0, 20, ..., 100},
yticklabels={$0$, $20$, $40$, $60$, $80$, $100$},
grid style=dashed,
ylabel={Exact Match (\%)},
xlabel={{Floating Point Precision}},
xlabel style={align=center,font=\small,yshift=0.2em},
ylabel style={font=\scriptsize,yshift=-1.2em},
y tick style={opacity=0},
y tick label style={font=\scriptsize},
x tick label style={font=\scriptsize},
ymajorgrids=true,
xmajorgrids=true,
tick align=inside,
legend pos=outer north east,
yticklabel style={/pgf/number format/precision=1,/pgf/number format/fixed zerofill},
xmin=0.5,
xmax=4.5,
ymin=-1,
ymax=105]
    \addplot[
        goob,mark=*,mark size=2pt,thick,mark options={fill=white,draw=goob,line width=1pt}
        ]
        coordinates {
        (1, 93.6)
        (2, 86.4)
        (3, 84.9)
        (4, 50.0)
        };
    \addplot[
        gooy,mark=*,mark size=2pt,thick,mark options={fill=white,draw=gooy,line width=1pt}
        ]
        coordinates {
        (1, 23.1)
        (2, 8.7)
        (3, 9.2)
        (4, 1.7)
        };
    \addplot[
        goor,mark=*,mark size=2pt,thick,mark options={fill=white,draw=goor,line width=1pt}
        ]
        coordinates {
        (1, 5.5)
        (2, 0.8)
        (3, 0.5)
        (4, 0.1)
        };

\end{axis}
}
\vspace{2cm}
\small{
\begin{axis}[
at={(9.0em,0)},
width=.27\textwidth, height=.22\textwidth ,
xtick={1,2,3,4},
xticklabels={$\scriptstyle \leq\!10^1$,$\scriptstyle \leq\!10^2$,$\scriptstyle \leq\!10^3$,$\scriptstyle \leq\!10^4$},
ytick={0, 20, ..., 100},
yticklabels={$0$, $20$, $40$, $60$, $80$, $100$},
grid style=dashed,
xlabel={{Integer Range}},
xlabel style={align=center,font=\small,yshift=0.2em},
y tick style={opacity=0},
y tick label style={font=\tiny},
ymajorgrids=true,
xmajorgrids=true,
yticklabels=\empty,
tick align=inside,
legend style={at={(-0.2,1.15)},anchor=south},
legend columns=3,
legend cell align={left},
xmin=0.5,
xmax=4.5,
ymin=-1,
ymax=105]

    \addplot[
        goob,mark=*,mark size=2pt,thick,mark options={fill=white,draw=goob,line width=1pt}
        ]
        coordinates {
        (1, 100)
        (2, 100)
        (3, 91.2)
        (4, 83.3)
        };
        \addlegendentry{\tiny{Two Operands}}

    \addplot[
        gooy,mark=*,mark size=2pt,thick,mark options={fill=white,draw=gooy,line width=1pt}
        ]
        coordinates {
        (1, 50.7)
        (2, 15.3)
        (3, 16.2)
        (4, 10.3)
        };
        \addlegendentry{\tiny{Three Operands}}

    \addplot[
        goor,mark=*,mark size=2pt,thick,mark options={fill=white,draw=goor,line width=1pt}
        ]
        coordinates {
        (1, 18.0)
        (2, 3)
        (3, 0.5)
        (4, 0.5)
        };
        \addlegendentry{\tiny{Four Operands}}
\end{axis}
}
\end{tikzpicture}
\vspace{-2mm}
\caption{Performance with different floating point precision (left) and integer range (right).}
\label{fig:synthesize_performance}
\end{wrapfigure}
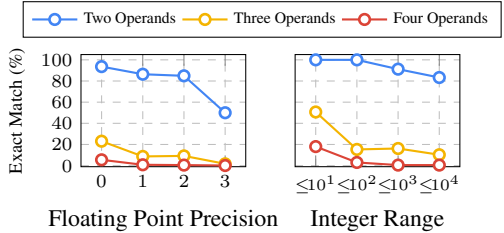

We sample a maximum of $50$ expressions for each different settings of complexity, and test these samples using large scale language model GPT-3~\citep{radford2019gpt3}. 
We conduct the study on GPT-3 in a few-shot manner: to unleash its potential, we pre-pend $10$ to $20$ expressions (having the same $f$, integer range, and floating point precision as the tested sample) together with the answers as the prompt. We then call the OpenAI API\footnote{\footnotesize{\url{https://openai.com/api}}} to get all the predictions, and evaluate the performance accordingly.

Results in Figure~\ref{fig:synthesize_performance} indicate that even the latest powerful GPT-3(\textit{Code-Davinci-002}) fails to achieve a satisfactory performance: 
(i) the prediction accuracy decreases largely as the number gets more complex, i.e., integer range or floating point precision of operands increases; (ii) the prediction accuracy also drops dramatically as the arithmetic relationship getting more complex, i.e., number of operands increases. 
In Appendix~\ref{sec:appendix-mathexp}, we also present the performance with our \ours framework, which is more robust to influence of floating point precision and integer range.

\section{Numerical Reasoning via Solving Linear Systems} \label{sec:synthesize}

The preliminary study demonstrates that the current language models are vulnerable to complex numbers.
For example, they have no chance to guess the answer to the sum of two floating point numbers with three decimal places.
However, the language model can perform reliably well when the operands are simple, i.e., relatively small integers.
Such observations motivate us to \textit{simplify the numbers before feeding them into language models}, thus enabling reliable neural-based numerical reasoning.
In this section, we first provide an overview of our framework \ours, and then we elaborate on each part of our framework in detail.

\subsection{Method Overview}

As mentioned above, our method can be integrated into language models in a plug-and-play manner at test time.
For the sake of clarification, in the following we refer to LMs that can steadily perform numerical reasoning as reasoning LMs.
They can be either LMs obtained by fine-tuning on datasets involving numerical reasoning, or LMs that perform numerical reasoning via in-context learning.

As shown in Figure~\ref{fig:model_intro}, our method generally involves three stages: (1) \textit{Operand Proposal}: given a paragraph, we first identify the numbers which are necessary for the reasoning LM to perform numerical reasoning (e.g., {\color{goor} $\mathbf{10,477}$}); (2) \textit{Number Substitution}: these proposed operands\footnote{We use the terms \textit{number} and \textit{operand} interchangeably.} are generally complex for language models, and thus they need to be substituted with randomly chosen simple numbers (e.g., {\color{goog} $\mathbf{10}$}) to make the model input simpler. Using the reasoning LM, we can obtain a set of predicted answers with respect to each substituted paragraph after several substitutions. (3) \textit{Arithmetic Relationship Inversion}: using these paragraphs and their answers as observed data, we can inversely derive the internal reasoning flow from the reasoning LM, i.e. the arithmetic expression between the operands (e.g., $y=x_1-x_2$). By applying the expression on the original numbers, the answer to the original paragraph can be obtained.

\subsection{Operand Proposal}\label{subsec:apply-ours}

There are often many numbers involved in a paragraph, and it is quite challenging to model the arithmetic relationships among all these numbers simultaneously.
Consequently, it is important during the operand proposal step to trim the prospective operands to a manageable size.
A straightforward strategy would be to select only the numbers pertinent to the answer as candidate operands, which is not trivial in practice since there is no intermediate supervision on the relevance between each number and the answer.

To address the issue, we provide a novel technique that employs number perturbation and the reasoning LM to measure the relevance systematically.
It is largely inspired by prior works that leverage an image classifier to quantify the relevance of pixels with image categories~\citep{samek2016evaluating} and its application on natural language tasks~\citep{liu-etal-2021-awakening}.
In their works, relevance is assessed by the degradation of the classifier score after erasing each pixel, where a substantial degradation indicates a strong relevance.
Similarly, we consider a number to be essential to the final answer if there is a difference between the model predictions before and after perturbing it.
Regarding perturbations, we implement it by adding a small adjustment to each number in the paragraph (e.g., $98.5 \rightarrow 98.6$) and evaluate whether the model prediction changes correspondingly.
Despite the fact that the reasoning LM hardly perform accurate calculations over numbers, we observe that LMs have strong context understanding capabilities about numbers and are sensitive to slight changes in the numbers used to forecast answer.
More details about the operand proposal mechanism can be found in Appendix~\ref{sec:operand}.

\subsection{Number Substitution}

After the operand proposal stage, a random set of numbers is generated to substitute the proposed operands sequentially.
These numbers are referred to as anchor numbers below.
Each anchor number is an integer between $1$ and $20$, a range that we believe reasoning LMs can easily handle.
Meanwhile, to minimize the effects of number substitution, we strive to maintain the order relationships among the numbers.
Taking the example from Figure~\ref{fig:model_intro}, we make the substitution number corresponding to $10,477$ larger than the one corresponding to $7,459$ since $10,477$ is larger than $7,459$.

Notably, the random number substitution must be repeated several times (e.g., three times in Figure~\ref{fig:model_intro}) to obtain a group of anchor numbers.
Along with the original question, each of these paragraphs is fed into the reasoning LM to predict the answer, which we call the anchor answer.
Typically, the number of anchor answers must exceed the number of operands for the subsequent arithmetic relationship inversion stage to be feasible.

\subsection{Arithmetic Relationship Inversion}
\label{sec:relation-inversion}

Given a collection of anchor numbers and anchor answers, the arithmetic relationship inversion stage investigates the relationship between these numbers and induces an expression to reflect it.
Taking the example from Figure~\ref{fig:model_intro}, a typical expression can be $y=x_1-x_2$, where $x_1$ and $x_2$ are both anchor numbers while $y$ is the anchor answer.

Although the example expression appears intuitive, deriving such an expression from data points is tremendously difficult because the solution space is theoretically infinite.
To make it practicable, as a first step, we begin by limiting the problem-solving space to compositions of binary operators, where each operator can be addition, subtraction, multiplication or division, the four most prevalent operators in numerical reasoning~\citep{dua2019drop}.
Meanwhile, there can be up to three compositions, which means the expression contains a maximum of four operands.
With such priors, the insoluble expression induction problem can be turned into a linear system solving problem, where the anchor numbers, the anchor answer, and their compositions constitute a linear system.
In this way, the problem of expression induction can be tackled by the \textit{solving algorithms} for linear systems, which will be elaborated in Section~\ref{sec:method}.
Finally, the answer can be reached in a trustworthy and interpretable manner by applying the derived expression to the original numbers.

\section{Solving Algorithm}\label{sec:method}

In this section, we introduce three algorithms that can derive expressions merely from anchor numbers and anchor answers, namely \textit{analytical-based}, \textit{search-based} and \textit{heuristic-based} algorithm.

\subsection{Formulation}
\label{sec:form}
Formally, given a paragraph and a question, we denote a group of anchor numbers as $\boldsymbol x=(x_1, x_2, \ldots, x_n)$ and the arithmetic relationship as an expression $f$, which should produce the answer $y$ by $y = f(\boldsymbol x)$.
The goal is to recover $f$ from different groups of anchor numbers $\mathbf X$ and corresponding anchor answers $\boldsymbol y$.
We propose to transform and formulate the arithmetic relationship inversion as solving a system of linear equations. Given expression $f(\boldsymbol x)$ with four fundamental arithmetic operations, we transform the equation $y = f(\boldsymbol x)$ by multiplying denominators on both sides when operator division exists, then we get:
\begin{equation}\label{eq:analytic}
    a_0 \cdot C + a_1 \cdot x_1 + a_2 \cdot x_2 + a_{3} \cdot y + a_4 \cdot x_1x_2 + \ldots + a_k \cdot (x_1x_2\ldots~x_ny) = 0
\end{equation}
For example, $y = 1 - x_1/x_2$ can be transformed to $x_2 - x_1 - x_2y = 0$.
Then uncovering $f(\boldsymbol x)$ is equivalent to solving $\boldsymbol a=(a_0, a_1, \ldots, a_k)$, which are coefficients of all possible polynomial basis combined by $x_1,,x_n$ and $y$, denoted as $\boldsymbol p$, where $k = 2^{n+1}-1$. Multiple groups of anchors $\mathbf X$ and $\boldsymbol y$ constitute multiple groups of values of polynomial basis, denoted as $\mathbf P$, then Equation~\ref{eq:analytic} can be denoted as $\mathbf P \boldsymbol a = \mathbf{0}$, which is a typical set of linear equations. 

\subsection{Analytical-based Algorithm}

To solve $\mathbf P \boldsymbol a = \mathbf{b}$, we can simply generate k+1 groups of anchor numbers as $\mathbf X$ and LMs' answers as $\boldsymbol y$, compute $\mathbf P$ based on $\mathbf X$ and $\boldsymbol y$, and finally get $\boldsymbol a = {{(\mathbf P)}^{-1} \boldsymbol b}$ when $\mathbf P$ is in full rank.
But notice that $y$ can be a linear weighted summation of $x_0, \ldots, x_n$ by itself, the coefficient matrix $\mathbf P$
may not be full-ranked. To address this, we generate $k$ groups of anchor numbers and add an additional constraint by setting $|\boldsymbol a| = \sum_{i=0}^{k}a_i = 1$. So we augment $\mathbf P$ with an all-one vector to $\mathbf P^*$ and finally get $\boldsymbol a = {{(\mP^*)}^{-1} \boldsymbol b}$, 
where $\boldsymbol b = (0, 0, \ldots, 0, 1)$. In practice, randomly sampled groups of anchor numbers can form a full-ranked $\mP^*$ with a very high probability, and one can even add a buffer by sampling a bit more groups of anchor numbers than $k$ to constitute different $\mP^*$s for cross validation.

The analytic method is theoretically complete to deduce arithmetic expressions in our pre-defined problem space. 
But in practice, LMs may produce incorrect results even for anchor numbers, especially when given a complex expression, so as to violate the analytic method which needs purely correct anchor answers. To best tolerate them, we then propose search-based and heuristic-based methods to better solve a noisy linear system. Gladly, the analytic method theoretical 
supports other methods in aspects such as guiding the number of anchors to sample to ensure a unique expression.

\subsection{Search-based Algorithm}

The search-based algorithm exhaustively explores the search space and finds out the most preferable arithmetic expression in the space. 
We constrain the search space  of $\boldsymbol a$ in Equation~\ref{eq:analytic} by: requiring $a_{1-n} \in \{-1, 0, 1\}$ for all coefficients of the non-constant terms, and for coefficient $a_0$ of constant term $C$, one can restrict the search range to a pre-defined set, e.g., $a_0 \in \{-100, -1, 0, 1, 100\}$ in our experiments for efficiency, and different from the analytic method that can easily solve constants in expressions. Constraints here mean that we only let this search algorithm cover $f(\boldsymbol x)$ with no more than one constant for efficiency. 
We then transform all searched polynomial-basis-based equations backwards into expressions because they have one-to-one mappings, e.g., from $x_2 - x_1 - x_2y = 0$ to $y= 1 - x_1 / x_2$.
We denote the space of expressions as $\mathbb{F}$, 
and for each $f_{i} \in \mathbb{F}$ and each group of anchor numbers $\mathbf X_j$ (using $m$ to denote the number of groups), we get $y_{ij}$ by applying $f_{i}$ to $\mathbf X_j$.

\setlength\intextsep{0pt}
\begin{wrapfigure}[16]{R}{0.45\textwidth}
    \centering
    \setlength{\tabcolsep}{0pt}
    \begin{minipage}{0.42\textwidth}
    \begin{algorithm}[H]
      \caption{\textsc{Search}}\label{alg:search}
      \begin{algorithmic}[1]
        \renewcommand{\algorithmicrequire}{\textbf{Input:}}
        \Require{parameters \( \mathbf{X}, \boldsymbol{\hat{y}}, \mathbb{F}, {c}_{threshold}\)}
        \renewcommand{\algorithmicensure}{\textbf{Output:}}
        \Ensure{Most preferable expression \( \widetilde{f} \)}
        \While {$j < m$} 
            \For{$f_{i} \in \mathbb{F}$}
                \State $y^*_{ij} \gets f_{i}(X_j)$
                \State $c_{i} \gets c_{i} + \mathbf{1}(y^*_{j} == \hat{y}_{ij})$
                \State $\epsilon_{i} \gets \epsilon_{i} + |y^*_j - \hat{y}_{ij}|$
            \EndFor
            \State $j \gets j + 1$
        \EndWhile
        \State $i^*_c \gets \arg \max \boldsymbol c, ~i^*_{\epsilon} \gets \arg \min \epsilon$
        \If {$c_{i^*} \geq c_{threshold}$}
            $\widetilde{f} \gets  f_{i^*_{c}}$
        \Else
            $\widetilde{f} \gets   f_{i^*_{\epsilon}}$
        \EndIf
      \end{algorithmic}
    \end{algorithm}
    \end{minipage}
\end{wrapfigure}
We define the prediction error between the target expression $\hat{f}$ and $f_{i}$ as $\epsilon (\hat{f}, f_i)$, which is calculated by $\epsilon(\hat{f}, f_i) = \sum_{j}\epsilon_{ij} = \sum_{j} \mathrm{abs}(\hat{y}_{j} - y_{ij})$, and the number of occurrence of exact matching as $\boldsymbol c_i$. We then find the most preferable expression with the minimum prediction error and the maximum number of exact matching. Specifically, when the number of exact matching exceeds a pre-defined $c_{threshold}$, we pick the expression $f_{i}$ with the highest $\boldsymbol c_i$; otherwise, we pick the expression $f_{i}$ with the lowest $\epsilon_i$.
The search process is sketched in Algorithm~\ref{alg:search}.

This method is robust for probably incorrect predictions, i.e., when model does not have sufficient number of exact matching, it is still capable to return the most nearest expression by selecting the one with the minimum error.
However, the search-based method can be challenged by exponentially explosive search space when the number of operands surges, and it's not efficient to search constant numbers that has a wide and even infinite range, neither.
\subsection{Heuristic-based Algorithm}

In this section, we introduce a heuristic-based algorithm, simulated annealing, which is efficient and does not need to search for the whole problem space, though it may produce sub-optimal results given a limited number of exploration steps. We follow the formulation introduced in Section~\ref{sec:form} and proposed a optimization target $\mathcal{L}_H$ to measure the L1 loss of $\mP\boldsymbol a$.
The pipeline includes:
(1) randomly initialize $\boldsymbol a$ with values \{-1, 0, 1\} and calculate initial $\mathcal{L}_H$; (2) randomly select $i$ from $0$ to $k$ and perturb $a_i$ by adding or subtracting a constant number (we use $1$ here);  (3) calculate new $\mathcal{L}_H$, and adopt the perturbation with a large probability if $\mathcal{L}_H$ decreases and with a low probability if it increases, balanced by a pre-defined temperature $T$, which decreases over steps; (4) return $\boldsymbol a$ if the number of steps is enough or $\mathcal{L}_H$ equals to zero, otherwise repeat from step 1. Note that, we restrict coefficients in $\boldsymbol a$ to be integers for simplicity, so different from the analytical method restricting $\sum_{i=0}^{k}a_i = 1$, we ensure only one of the coefficients of $y$-related polynomial basis \{$y$, $x_1y$, \ldots, $x_1x_2\ldots~x_ny$\} to be non-zero (with a static value $1$) and at least two coefficients in $\boldsymbol a$ are non-zero during the whole initialization and perturbation process to avoid some infeasible local optimal.

\begin{wraptable}[8]{R}{0.48\textwidth}
    \centering
    \setlength{\tabcolsep}{0pt}
    \begin{minipage}{0.48\textwidth}
    {
    \caption{Comparison of solving algorithms.} \label{tab:method-comparison}
    }  \begin{tabular}{lccc}
            \toprule
                & Optimum~~~       & Robustness~~~    & Scalability   \\
            \midrule
            Analytical    & \cmark    & \xmark      & \cmark \\
            Search        & \cmark    & \cmark    & \xmark   \\
            Heuristic     & \xmark      & \cmark    & \cmark \\
            \bottomrule
        \end{tabular}
    \end{minipage}
\end{wraptable}

In summary, Table \ref{tab:method-comparison} shows the strong and weak points of these algorithms. In the problem space introduced in Section~\ref{sec:relation-inversion} within at most four operands, the search-based method does not have scalability issues, so it achieves best performance in our experiments because it's robust to LMs' predictions and can retrieve optimal expression through exhaustive search  except rare constants.

\section{Experiments}\label{sec:exp}

In this section, we integrate \ours with various language models as backbones and evaluate the effectiveness of \ours on two well-known numerical reasoning benchmarks.

\subsection{Experimental Setup}

\noindent\textbf{Datasets}
We perform experiments on DROP~\citep{dua2019drop}, AddSub and MultiArith, of which the latter two are widely used subsets from MAWPS~\citep{roy2015solving}.
DROP is a reading comprehension benchmark that focuses on numerical reasoning and has a variety of answer types, including \textit{span}, \textit{number}, and \textit{date}.
The experimental results of DROP are evaluated with the official evaluation metrics Exact Match (EM) and F1.
As for MAWPS, it consists of math word problems which also require numerical reasoning ability.
The subset AddSub features relatively easier numerical reasoning, whereas MultiArith necessitates multi-step numerical calculations.
The EM metric is used to evaluate the results of AddSub and MultiArith.
More details can be found in Appendix~\ref{sec:appendix-exp}.

\noindent\textbf{Backbone and Baselines} 
On DROP, we adopt two kinds of LMs as our backbones, including (i) Vanilla LMs: BART~\citep{lewis2020bart} and T5~\citep{raffel2020t5}, (ii) Reasoning LMs: \tapex~\citep{liu2021tapex} and \poet~\citep{pi2022reasoning}.

We also compare the performance of our method with previous specialized models designed for the DROP dataset, such as NumNet~\citep{ran2019numnet}, NeRd~\citep{chen2020neural}, MTMSN~\citep{hu2019mtmsn} and QDGAT~\citep{chen2020qdgat}.
All models are fine-tuned on the DROP train set, and the best validation set performance is reported.
If no explicit declarations are included, all LMs here are of large size and contain approximately $350$M of parameters.
On AddSub and MultiArith, we adopt GPT-3 Code-Davinci-002 (GPT-3) ~\citep{radford2019gpt3} with different prompting techniques as our backbones: Chain-of-Thought Prompting (Chain)~\citep{wei2022chain} and the Zero-shot Chain-of-Thought Prompting (Zero-shot Chain)~\citep{kojima2022large}.
We compare our results to the PaLM model~\citep{chowdhery2022palm}.
Unless otherwise specified, these models perform numerical reasoning by zero-shot/few-shot in-context learning, and the few-shot demonstrations are the $8$ samples released by \citet{wei2022chain}.

\subsection{Implementation Details}

\noindent\textbf{Hyperparameter Selection}
For fine-tuning, we apply Adam~\citep{DBLP:conf/iclr/LoshchilovH19} optimizer. The fine-tuning epochs are set as 50. For BART models (i.e., BART and \poet-SQL), we follow previous works~\citep{pi2022reasoning} to set the batch size as $128$ and the learning rate as $3\times10^{-5}$.
For T5, we decrease the batch size to $32$ due to the computational budget. The early stop technique is used to save training time.
For GPT-3 API, we keep the temperature as default setting $0$, and set the maximum output tokens to $128$.
As for anchor number groups: the group size is $6/8/10$ corresponding to corresponding to $2/3/4$ operands on DROP; the group size is $4$ on AddSub, and $10$ on MultiArith because MultiArith requires more compositional operations.
More details can be found in Appendix~\ref{sec:appendix-exp}.

\noindent\textbf{Design Choices on DROP} 
Following previous work, we apply two general-purpose numerical designs on the DROP dataset.
First, we employ the character-level rather than subword-level number representation, which proves to be more effective~\citep{wallace2019nlp,pi2022reasoning}. 
Second, we employ the reverse decoding technique, which proves to be a successful design to mimic arithmetic carry~\citep{geva2020injecting}.
Meanwhile, as mentioned above, the search-based algorithm has difficulties in covering expressions including constants.
Considering the constant $100$ is frequently used for percentage calculations (e.g., ``How many percent of the national population does not live in Bangkok?''), we add it to be one candidate in DROP.

\subsection{Experimental Results}

\begin{table}[tb]
\centering
\small
\caption{Experimental results of \ours \textit{w.} various solving algorithms on the DROP numeric subset.}
\label{tab:drop-for-numeracy}
\scalebox{1.0}{
\begin{tabular}{ccll}
\toprule
\textbf{LM} & \textbf{Algorithm} & \textbf{F1(\%) on Hard} & {\textbf{F1(\%) on Total}} \\ 
\midrule
\multirow{4}{*}{BART}   & -- &  $30.4$  & $66.4$\\
                        &  Analytical   & $46.4$ {\footnotesize $(+16.0)$}  & $69.3$ {\footnotesize $(+2.9)$} \\
                        &  Search  & $64.8$ {\footnotesize $(+30.4)$}  & $75.2$ {\footnotesize $(+8.8)$} \\
                        &  Heuristic & $52.8$ {\footnotesize $(+22.4)$}  & $71.7$ {\footnotesize $(+5.3)$} \\
\midrule
\multirow{4}{*}{\poet-SQL}      &  -- &  $66.8$  & $78.4$\\
                                &  Analytical   & $73.3$ {\footnotesize $(+6.5)$}  & $80.0$ {\footnotesize $(+1.6)$}    \\ 
                                &  Search  & $76.9$ {\footnotesize $(+10.1)$}  &  $81.4$ {\footnotesize $(+3.0)$} \\
                                &  Heuristic & $73.0$ {\footnotesize $(+6.2)$} &   $80.5$ {\footnotesize $(+2.1)$}\\
\bottomrule
\end{tabular}%
}
\end{table}

\begin{table}[tb]
\centering
\small
\caption{Experimental results on the validation set of DROP dataset.}
\label{tab:drop-for-bart}
\scalebox{1.0}{%
    \begin{tabular}{lll}
    \toprule
    \textbf{Models}                               & \textbf{EM(\%)}                        & \textbf{F1(\%)}     \\ \midrule
    \multicolumn{3}{c}{\textit{Specialized Models}} \\
    NumNet~\citep{ran2019numnet}                         & $64.9$ & $68.3$ \\
    MTMSN~\citep{hu2019mtmsn}                           & $76.7$ & $80.5$ \\
    NeRd~\citep{chen2020neural}                    & $78.6$ & $81.9$ \\
    QDGAT~\citep{chen2020qdgat}                           & ${84.1}$ & ${87.1}$ \\
    \multicolumn{3}{c}{\textit{Vanilla LMs}}                \\
    BART~\citep{lewis2020bart}                     &   $67.4$   &   $70.6$   \\ 
    ~~~~\textit{w.} \ours            &   $72.9$ {\footnotesize $(+5.5)$}   &   $76.1 $ {\footnotesize$(+5.5)$}    \\
    T5~\citep{raffel2020t5}                        &   $61.0$        &   $64.6$        \\ 
    ~~~~\textit{w.} \ours            &  $69.9$ {\footnotesize $(+8.9)$}         &    $73.5$ {\footnotesize $(+8.9)$}       \\
    \multicolumn{3}{c}{\textit{Reasoning LMs}}               \\
\tapex~\citep{liu2021tapex}           &   $76.3$   &   $79.3$ \\
    ~~~~\textit{w.} \ours            &   $78.5$ {\footnotesize $(+2.2)$}   &   $81.6$ {\footnotesize $(+2.3)$} \\
    \poet-SQL~\citep{pi2022reasoning}          &   $76.9$   & $80.0$ \\
    ~~~~\textit{w.} \ours            &   $78.2$ {\footnotesize $(+1.3)$}  &    $82.0$ {\footnotesize $(+2.0)$} \\ 
    \bottomrule
    \end{tabular}%
}
\end{table}

Since our work focuses on addressing arithmetic problems, we first evaluate suggested solving algorithms via their performance on the DROP subset whose answers are numbers (i.e., numeric subset).
Meanwhile, we select cases in which the answer is greater than $1000$, identify them as ``hard'' cases, and additionally report the average performance on them.
As shown in Table~\ref{tab:drop-for-numeracy}, all of our proposed algorithms significantly improve the performance of LMs, especially in hard cases.
For example, the search-based algorithm boosts BART with an absolute $30.4\%$ improvement on hard cases.
The full results of the performance comparison can be found in Appendix~\ref{sec:full-results}.
Notably, since the search-based algorithm is the most effective, we apply it as the default algorithm in \ours.

\begin{table}[tb]
    \centering
    \small
    \caption{Experimental results of different methods on AddSub and MultiArith.}
    \label{tab:codex-few-shot}
    \scalebox{1.0}{
    \begin{tabular}{l p{5cm} p{2.0cm}p{2.0cm}}
        \toprule
        \multicolumn{1}{l}{\textbf{Language Model}} & \textbf{Setting}         & {\textbf{AddSub}} & {\textbf{MultiArith}}\\
        \midrule
        \multirow{2}{*}{PaLM (540B)} 
                & Standard~\citep{chowdhery2022palm}  & $-$    & $42.2$ \\
                & Chain~\citep{wei2022chain}     & $91.9$ & $94.7$ \\
        \midrule
        \multirow{4}{*}{GPT-3 (175B)}       
                & Zero-shot Chain~\citep{kojima2022large}             & $66.6$ & $63.8$\\
                & ~~~~\textit{w.} {\footnotesize \ours}   & $89.4$ {\footnotesize $(+22.8)$}  & $80.0$ {\footnotesize $(+16.2)$} \\
                & Chain~\citep{wei2022chain}                     & $88.4$    & $96.7$ \\
                &  ~~~~\textit{w.} {\footnotesize \ours}   & $90.9$ {\footnotesize $(+2.5)$}   & $98.7$ {\footnotesize $(+2.0)$}\\
        \bottomrule
    \end{tabular}
    }
\end{table}

Table~\ref{tab:drop-for-bart} shows the experimental results of different models on DROP dataset.
As shown, \ours can bring consistent and significant improvements over all backbone LMs, especially for the vanilla LMs.
Taking the T5 model as an example, it could be boosted by a maximum of $8.9\%$ with \ours.
Even for \poet-SQL which are already pre-trained for numerical reasoning, our method yields a $2.0\%$ F1 improvement, pushing the best LM performance to $82.0\%$ F1.
Table~\ref{tab:codex-few-shot} presents the experimental results on AddSub and MultiArith.
The results indicate that our approach is surprisingly effective for giant LMs and can further boost the performance of chain-of-thought prompting.

\section{Model Analysis}

\noindent\textbf{Arithmetic Relationship Inversion}
In addition to performance improvement, \ours features the ability to derive an arithmetic expression for each question, whereas no such information is available during training.
To better understand if these expressions align with question intentions, we collect all derived expressions and categorize them into four types in Table~\ref{tab:empirical-study}.
As demonstrated, the majority of expressions contain addition and subtraction between variables and constants, which are largely consistent with the question intention, highlighting the superior interpretability of \ours.

\begin{table}[t]
\centering
\small
\caption{Case study on derived expressions using \poet-SQL \textit{w.} \ours on DROP. Listed are, the intention, the example question with intention trigger words (i.e., the {\color{NavyBlue} \bf colorful} spans) and the derived expression, and the proportion of each intention.}
\label{tab:empirical-study}
\begin{tabular}{c p{8.3cm} c}
\toprule
\bf Question Intention & \bf Example Question with [Derived Expression] & \bf Proportion \\
\midrule
{\color{NavyBlue} \bf Addition}      & How many {\color{NavyBlue} \bf total} yards of touchdown passes were there? {\color{NavyBlue}~$\mathtt{[y=x_1+x_2+x_3]}$} & $8.92\%$\\
\midrule
{\color{Orange} \bf Diff Constant}    & How many {\color{Orange} \bf in percent} in the county from the census of 2000 {\color{Orange} \bf weren't} English? {\color{Orange}$\mathtt{[y=100-x]}$}   & $36.49\%$\\
\midrule
{\color{Plum} \bf Subtraction}  &  How many {\color{Plum} \bf more} percentages of people were germans {\color{Plum} \bf compared to} irish? {\color{Plum}$\mathtt{[y=x_1-x_2]}$}& $54.25\%$\\
\midrule
{\color{Green} \bf Composition}             & How many {\color{Green} \bf more} Albanian citizens were there {\color{Green} \bf compared to} Bulgarian and Georgia citizens {\color{Green} \bf combined ? } {\color{Green}$\mathtt{[y=x_0-(x_1+x_2)]}$} & $0.34\%$\\
\bottomrule
\end{tabular}
\end{table}

\noindent\textbf{Solving Algorithm Robustness}
The possibility that the anchor answers provided by reasoning LMs are inaccurate presents a challenge for the solving algorithms.
To measure the robustness of our solving algorithms, we roughly decrease the probability that anchor answers are correct by decreasing the number of few-shot demonstrations in Figure~\ref{fig:analysis-few-shot}.
As shown, even though the backbone LM performance drops to $60.0\%$, the improvement of \ours is still as high as to $5.1\%$, suggesting its robustness.

\noindent\textbf{Number Substitution}
To study the impact of different factors during the number substitution stage, we conduct experiments on \ourtask in Figure~\ref{fig:ablation_study_range_time}.
As demonstrated, expanding the range of anchor numbers results in a minor performance drop, showing that the reasoning LM is more familiar with small integers.
Furthermore, increasing the size of anchor number groups gives a large improvement on the performance, especially when there are four operands.

\noindent\textbf{Limitation Discussion}
The first limitation of our framework is that we cannot support expressions that cannot be solved with linear systems.
For example, with respect to the question ``How many yards was Donovan McNabb's longest rushing TD?'', the expected expression $\mathtt{[y=\max_i(x_i)]}$ is not supported by \ours.
Second, the framework is less efficient when there are many operands.
On the one hand, the group of anchor numbers would be quite huge, making the algorithm's runtime unacceptable.
For example, when expanding to $5$ operands, number substitution must be performed at least $50$ times.
On the other hand, for the search-based algorithm, the search space will increase exponentially, making the algorithm impracticable.
Last, we assume a certain level of numeracy understanding of the reasoning LM.
Therefore, if the reasoning LM is unable to comprehend the numeracy relationship, our method would not work well.

\pgfplotstableread[row sep=\\,col sep=&]{
    Scale & codex & ours \\
    4  & 60.0 & 65.1 \\
    3 & 84.0 & 87.6 \\
    2  & 84.5 & 88.4 \\
    1  & 88.4 & 90.9  \\
}\datatable

\begin{figure}[t]
\begin{minipage}{0.44\linewidth}
 \centering
  \begin{tikzpicture}[scale=0.8]
        \begin{axis}[
                ybar,
                enlarge x limits=0.2,
                width=1.1\textwidth,
                height=0.63\textwidth,
                bar width=1.2em,
                legend style={at={(0.5,1.35)},
                anchor=north,legend columns=-1},
                ylabel style={align=center,yshift=-1em},
                ytick={50,60,...,100},
                y tick style={opacity=0},
                xtick={1,2,3,4},
                xticklabels={$8$, $4$, $2$, $1$},
                ymajorgrids=true,
                xmajorgrids=true,
                grid style=dashed,
                nodes near coords,
                y tick label style={font=\small},
                every node near coord/.append style={font=\scriptsize,color=black},
                nodes near coords align={vertical},
                tick align=inside,
                xmax=4,
                xmin=1,
                ymin=50,ymax=100,
                ylabel={Exact Match (\%)},
                xlabel={\# of Few-shot Examples},
                xlabel style={align=center,yshift=0.2em}
            ]
            \addplot +[fill=goor!90,draw=goor] table[x=Scale,y=codex]{\datatable};
            \addplot +[fill=gooy!90,draw=gooy] table[x=Scale,y=ours]{\datatable};
            
            \legend{\small{Chain}, \small{Chain \textit{w.} \ours}}
        \end{axis}
    \end{tikzpicture}
    \vspace{-2mm}
 \caption{The experimental results of Chain and Chain \textit{w.} \ours on AddSub as the number of few-shot examples decreases.}
\label{fig:analysis-few-shot}
\end{minipage}
 \hspace{0.1cm}
\begin{minipage}{0.55\linewidth}
 \centering
 \begin{tikzpicture}[scale=0.8]
\small{
\begin{axis}[
at={(0,0)},
width=.66\textwidth, height=.5\textwidth ,
xtick={1,2,3,4},
xticklabels={$\leq5$, $\leq10$, $\leq20$, $\leq100$},
ytick={0, 20, ..., 100},
yticklabels={$0$, $20$, $40$, $60$, $80$, $100$},
grid style=dashed,
ylabel={Exact Match (\%)},
xlabel={Anchor Number Range},
xlabel style={align=center,font=\small,yshift=0.2em},
ylabel style={font=\small,yshift=-1.2em},
y tick style={opacity=0},
y tick label style={font=\small},
ymajorgrids=true,
xmajorgrids=true,
tick align=inside,
yticklabel style={/pgf/number format/precision=1,/pgf/number format/fixed zerofill},
xmin=0.5,
xmax=4.5,
ymin=-1,
ymax=110]
\addplot[
    goob,mark=*,mark size=2pt,thick,mark options={fill=white,draw=goob,line width=1pt}
    ]
    coordinates {
    (1, 100)
    (2, 100)
    (3, 100)
    (4, 100)
    };
\addplot[
    gooy,mark=*,mark size=2pt,thick,mark options={fill=white,draw=gooy,line width=1pt}
    ]
    coordinates {
    (1, 96)
    (2, 90)
    (3, 88)
    (4, 72)
    };
\addplot[
    goor,mark=*,mark size=2pt,thick,mark options={fill=white,draw=goor,line width=1pt}
    ]
    coordinates {
    (1, 32)
    (2, 26)
    (3, 28)
    (4, 24)
    };

\end{axis}
}
\vspace{1cm}
\small{
\begin{axis}[
at={(13.0em,0)},
width=.7\textwidth, height=.5\textwidth ,
xtick={1,2,3,4,5,6},
xticklabels={$5$, $10$, $15$, $20$, $25$, $30$},
ytick={0, 20, ..., 100},
yticklabels={$0$, $20$, $40$, $60$, $80$, $100$},
grid style=dashed,
xlabel={Anchor Number Group Size},
xlabel style={align=center,font=\small,yshift=0.2em},
y tick style={opacity=0},
ymajorgrids=true,
xmajorgrids=true,
yticklabels=\empty,
tick align=inside,
legend style={at={(-0.05,1.12)},anchor=south},
legend columns=3,
legend cell align={left},
xmin=0.5,
xmax=6.5,
ymin=-1,
ymax=110]

\addplot[
    goob,mark=*,mark size=2pt,thick,mark options={fill=white,draw=goob,line width=1pt}
    ]
    coordinates {
    (1, 100)
    (2, 100)
    (3, 100)
    (4, 100)
    (5, 100)
    (6, 100)
    };
    \addlegendentry{\small{Two Operands}}

\addplot[
    gooy,mark=*,mark size=2pt,thick,mark options={fill=white,draw=gooy,line width=1pt}
    ]
    coordinates {
    (1, 96)
    (2, 98)
    (3, 100)
    (4, 100)
    (5, 100)
    (6, 100)
    };
    \addlegendentry{\small{Three Operands}}

\addplot[
    goor,mark=*,mark size=2pt,thick,mark options={fill=white,draw=goor,line width=1pt}
    ]
    coordinates {
    (1, 32)
    (2, 64)
    (3, 66)
    (4, 70)
    (5, 84)
    (6, 84)
    };
    \addlegendentry{\small{Four Operands}}
\end{axis}
}
\end{tikzpicture}
\vspace{-2mm}
     \caption{The experimental results of \ours on \ourtask with different choices of anchor number range (\textbf{left}) and anchor number groups (\textbf{right}).}
     \label{fig:ablation_study_range_time}
 \end{minipage}
 \vspace{-3mm}
\end{figure}

\section{Related Work}

\noindent\textbf{Numerical Reasoning via Specialized Models} 
Previous works generally design trainable specialized modules and equip LMs with them to tackle different kinds of numerical reasoning problems (e.g., counting).
While these methods work well on specific datasets~\citep{dua2019drop, andor2019giving, hu2019mtmsn, ding2019cognitive}, they are hardly suited across different datasets and backbone LMs~\citep{chen2020neural}.
Differently, since our method does not require additional model training, it is applicable to almost all models, even those that only provide an inference interface (e.g., GPT-3).
As for methods that first generate programs or logic forms, it is quite laborious to define domain-specific language and collect corresponding training data \citep{berant-etal-2013-semantic}.
Unlike them, our methods does not require extra annotated programs.
Instead, our method allows for the program discovery from examples via solving linear systems.

\noindent\textbf{Numerical Reasoning via Pre-training}
This line of work always focuses on the pre-training of language models with corpus which involves reasoning.
The corpus can be reasoning-oriented natural language texts from Internet~\citep{deng-etal-2021-reasonbert,lewkowycz2022solving}, human-designed templates filled by different data sources~\citep{geva2020injecting,yoran2022turning}, or programs with rich reasoning semantics~\citep{liu2021tapex,pi2022reasoning}.
Although this kind of pre-training allows language models to perform better reasoning, they still require considerable computation budgets during pre-training and may still be challenged by complex numbers. 
In contrast, our method is efficient since it can be integrated into existing models without further training or pre-training. 

\noindent\textbf{Numerical Reasoning in Giant Language Models}
Recent works demonstrate that with proper prompting, giant language models (e.g., GPT-3) perform much better than smaller ones on several reasoning tasks~\citep{wei2022chain,Wei2022EmergentAO,kojima2022large,li2022advance, zhou2022least,wang2022self}.
For example, with the chain-of-thought prompting, the few-shot PaLM model~\citep{chowdhery2022palm} can beat the previous best fine-tuned model on math word problems.
However, their conclusions do not generalize to non-giant language models.
Different from them, our method can be simultaneously applied to language models ranging from millions (e.g., BART) to billions (e.g., GPT-3). 
Moreover, our work is orthogonal to these giant LMs and can be complementary to each other.
For example, Section~\ref{sec:exp} shows that our approach can further boost the numerical reasoning capability of GPT-3 with chain-of-thought prompting.

\section{Conclusion}

In this work, we present \ours, a framework which can elicit numerical reasoning in language models at test time.
Motivated by the fact that language models usually excel at simple numbers, \ours uses simple numbers as anchors to inversely derive the implicitly inferred arithmetic expressions from language models, and subsequently apply these expressions to complex numbers to perform numerical reasoning.
With modeling the expression derivation as solving linear systems, we propose three kinds of algorithms to achieve \ours with noisy signals.
Experimental results on several numerical reasoning benchmarks demonstrate that \ours can be integrated to a variety of language models, and can greatly improve their performance under zero-shot, few-shot, and fine-tuning scenarios.
Our work provides a new perspective towards tackling numerical reasoning, which can be potentially applied to more language models and numerical reasoning tasks.

\section{Acknowledgements}

We would like to thank Xuanyu Dong, who is working at Harvest Fund, for helping us with the linearization method based on polynomial basis from an analytical and mathematical perspective.


\bibliography{iclr2023_conference}
\bibliographystyle{iclr2023_conference}

\newpage
\appendix

\section{Preliminary Study Details}\label{sec:appendix-mathexp}

Here we present the model performance on \ourtask of GPT-3 with different solving algorithms in Figure~\ref{fig:searc_algo} and Figure~\ref{fig:ana_algo}.
We can conclude that: (1) both algorithms are not sensitive with either the floating point precision or the integer range; (2) the search-based algorithm is most robust than the analytical-based algorithm with respect to the number of operands.

\vspace{30pt}
\begin{figure}[h]
\centering
\begin{minipage}{0.48\textwidth}
    \begin{figure}[H]
    \centering
    \begin{tikzpicture}
    \small{
    \begin{axis}[
    at={(0,0)},
    width=.55\textwidth, height=.55\textwidth ,
    xtick={1,2,3,4},
    xticklabels={$0$, $1$, $2$, $3$},
    ytick={0, 20, ..., 100},
    yticklabels={$0$, $20$, $40$, $60$, $80$, $100$},
    grid style=dashed,
    ylabel={Exact Match (\%)},
    xlabel={{Floating Point Precision}},
    xlabel style={align=center,font=\small,yshift=0.2em},
    ylabel style={font=\scriptsize,yshift=-1.2em},
    y tick style={opacity=0},
    y tick label style={font=\scriptsize},
    x tick label style={font=\scriptsize},
    ymajorgrids=true,
    xmajorgrids=true,
    tick align=inside,
    legend pos=outer north east,
    yticklabel style={/pgf/number format/precision=1,/pgf/number format/fixed zerofill},
    xmin=0.5,
    xmax=4.5,
    ymin=-1,
    ymax=105]
        \addplot[
            goob,mark=*,mark size=2pt,thick,mark options={fill=white,draw=goob,line width=1pt}
            ]
            coordinates {
            (1, 100)
            (2, 100)
            (3, 100)
            (4, 100)
            };
        \addplot[
            gooy,mark=*,mark size=2pt,thick,mark options={fill=white,draw=gooy,line width=1pt}
            ]
            coordinates {
            (1, 96)
            (2, 96)
            (3, 96)
            (4, 96)
            };
        \addplot[
            goor,mark=*,mark size=2pt,thick,mark options={fill=white,draw=goor,line width=1pt}
            ]
            coordinates {
            (1, 32)
            (2, 32)
            (3, 32)
            (4, 32)
            };
    
    \end{axis}
    }
    \vspace{2cm}
    \small{
    \begin{axis}[
    at={(9.0em,0)},
    width=.55\textwidth, height=.55\textwidth ,
    xtick={1,2,3,4},
    xticklabels={$\scriptstyle \leq\!10^1$,$\scriptstyle \leq\!10^2$,$\scriptstyle \leq\!10^3$,$\scriptstyle \leq\!10^4$},
    ytick={0, 20, ..., 100},
    yticklabels={$0$, $20$, $40$, $60$, $80$, $100$},
    grid style=dashed,
    xlabel={{Integer Range}},
    xlabel style={align=center,font=\small,yshift=0.2em},
    y tick style={opacity=0},
    y tick label style={font=\tiny},
    ymajorgrids=true,
    xmajorgrids=true,
    yticklabels=\empty,
    tick align=inside,
    legend style={at={(-0.2,1.15)},anchor=south},
    legend columns=3,
    legend cell align={left},
    xmin=0.5,
    xmax=4.5,
    ymin=-1,
    ymax=105]
    
        \addplot[
            goob,mark=*,mark size=2pt,thick,mark options={fill=white,draw=goob,line width=1pt}
            ]
            coordinates {
            (1, 100)
            (2, 100)
            (3, 100)
            (4, 100)
            };
            \addlegendentry{\tiny{Two Operands}}
    
        \addplot[
            gooy,mark=*,mark size=2pt,thick,mark options={fill=white,draw=gooy,line width=1pt}
            ]
            coordinates {
            (1, 96)
            (2, 96)
            (3, 96)
            (4, 96)
            };
            \addlegendentry{\tiny{Three Operands}}
    
        \addplot[
            goor,mark=*,mark size=2pt,thick,mark options={fill=white,draw=goor,line width=1pt}
            ]
            coordinates {
            (1, 32)
            (2, 32)
            (3, 32)
            (4, 32)
            };
            \addlegendentry{\tiny{Four Operands}}
    \end{axis}
    }
    \end{tikzpicture}
    \caption{Performance over different floating point precision (left) and integer range (right) on \ourtask of GPT-3 \textit{w.} search-based algorithm. }
    \label{fig:searc_algo}
    \end{figure}
\end{minipage}\hfill
\begin{minipage}{0.5\textwidth}

    \begin{figure}[H]
    \centering
    \begin{tikzpicture}
    \small{
    \begin{axis}[
    at={(0,0)},
    width=.55\textwidth, height=.55\textwidth ,
    xtick={1,2,3,4},
    xticklabels={$0$, $1$, $2$, $3$},
    ytick={0, 20, ..., 100},
    yticklabels={$0$, $20$, $40$, $60$, $80$, $100$},
    grid style=dashed,
    ylabel={Exact Match (\%)},
    xlabel={{Floating Point Precision}},
    xlabel style={align=center,font=\small,yshift=0.2em},
    ylabel style={font=\scriptsize,yshift=-1.2em},
    y tick style={opacity=0},
    y tick label style={font=\scriptsize},
    x tick label style={font=\scriptsize},
    ymajorgrids=true,
    xmajorgrids=true,
    tick align=inside,
    legend pos=outer north east,
    yticklabel style={/pgf/number format/precision=1,/pgf/number format/fixed zerofill},
    xmin=0.5,
    xmax=4.5,
    ymin=-1,
    ymax=105]
        \addplot[
            goob,mark=*,mark size=2pt,thick,mark options={fill=white,draw=goob,line width=1pt}
            ]
            coordinates {
            (1, 100)
            (2, 100)
            (3, 100)
            (4, 100)
            };
        \addplot[
            gooy,mark=*,mark size=2pt,thick,mark options={fill=white,draw=gooy,line width=1pt}
            ]
            coordinates {
            (1, 48.4)
            (2, 48.4)
            (3, 48.4)
            (4, 48.4)
            };
        \addplot[
            goor,mark=*,mark size=2pt,thick,mark options={fill=white,draw=goor,line width=1pt}
            ]
            coordinates {
            (1, 5.12)
            (2, 5.12)
            (3, 5.12)
            (4, 5.12)
            };
    
    \end{axis}
    }
    \vspace{2cm}
    \small{
    \begin{axis}[
    at={(9.0em,0)},
    width=.55\textwidth, height=.55\textwidth ,
    xtick={1,2,3,4},
    xticklabels={$\scriptstyle \leq\!10^1$,$\scriptstyle \leq\!10^2$,$\scriptstyle \leq\!10^3$,$\scriptstyle \leq\!10^4$},
    ytick={0, 20, ..., 100},
    yticklabels={$0$, $20$, $40$, $60$, $80$, $100$},
    grid style=dashed,
    xlabel={{Integer Range}},
    xlabel style={align=center,font=\small,yshift=0.2em},
    y tick style={opacity=0},
    y tick label style={font=\tiny},
    ymajorgrids=true,
    xmajorgrids=true,
    yticklabels=\empty,
    tick align=inside,
    legend style={at={(-0.2,1.15)},anchor=south},
    legend columns=3,
    legend cell align={left},
    xmin=0.5,
    xmax=4.5,
    ymin=-1,
    ymax=105]
    
        \addplot[
            goob,mark=*,mark size=2pt,thick,mark options={fill=white,draw=goob,line width=1pt}
            ]
            coordinates {
            (1, 100)
            (2, 100)
            (3, 100)
            (4, 100)
            };
            \addlegendentry{\tiny{Two Operands}}
    
        \addplot[
            gooy,mark=*,mark size=2pt,thick,mark options={fill=white,draw=gooy,line width=1pt}
            ]
            coordinates {
            (1, 48.4)
            (2, 48.4)
            (3, 48.4)
            (4, 48.4)
            };
            \addlegendentry{\tiny{Three Operands}}
    
        \addplot[
            goor,mark=*,mark size=2pt,thick,mark options={fill=white,draw=goor,line width=1pt}
            ]
            coordinates {
            (1, 5.2)
            (2, 5.2)
            (3, 5.2)
            (4, 5.2)
            };
            \addlegendentry{\tiny{Four Operands}}
    \end{axis}
    }
    \end{tikzpicture}
    \caption{Performance over different floating point precision (left) and integer range (right) on \ourtask of GPT-3 \textit{w.} analytical-based algorithm. }
    \label{fig:ana_algo}
    \end{figure}
\end{minipage}\hfill
\end{figure}

\section{Operand Proposal Details}\label{sec:operand}

In Section~\ref{subsec:apply-ours}, we mention that the textual context on a realistic dataset may be noisy, i.e., contains irrelevant numbers, thus we need to locate the operand number first. We substitute $10$ times for each number appearing in the paragraph, if the output gives $\geq 3$ different prediction numbers out of $10$, we decide the current tested number is involved to the answer. Moreover, we substitute numbers following a template: suppose the original number $x$ is with precision $\mathtt{p}$, then the substituted numbers can be represented as $x + k \cdot 10^{\mathtt{p}}$, where $k \in \{-5, -4, -3, -2, -1, 1, 2, 3, 4, 5\}$.

\section{Experiments}\label{sec:appendix-exp}

\subsection{Experimental Setup}
\vspace{20pt}
\begin{table}[h]
  \centering
  \caption{Statistics of DROP dataset}
  \label{tab:dataset-data}
  \scalebox{1.0}{
    \begin{tabular}{lcccc}
    \toprule
    \multicolumn{1}{c}{\multirow{2}[4]{*}{\bf Dataset}} & \multicolumn{2}{c}{\bf Train} & \multicolumn{2}{c}{\bf Dev}
    \\
         \cmidrule(lr){2-3}
     \cmidrule(lr){4-5}
          & {\# Questions} & {\# Docs} & {\# Questions} & {\# Docs} \\
    \midrule
    DROP & $77,409$  & $5,565$  &  $9,536$ & $582$ \\
    \bottomrule
    \end{tabular}
    }
\end{table}
\vspace{30pt}

\begin{table}[h]
  \centering
  \caption{Statistics of MAWPS dataset}
  \label{tab:mawps-data}
  \scalebox{0.9}{
    \begin{tabular}{lc}
    \toprule
    {\bf Subset} & {\bf \# Questions}\\
    \midrule
    AddSub   & $395$ \\
    MultiArith & $600$ \\
    \bottomrule
    \end{tabular}
    }
\end{table}

\vspace{20pt}
For BART, we implement the fine-tuning methods using the Huggingface transformers library~\citep{wolf2020transformers} on $4$ V100 16GB GPUs.
We use \texttt{BART}$_\text{LARGE}$\citep{lewis2020bart} as our backbone. We use same-scale reasoning-pretrained \poet-SQL and \tapex\ models in experiments.
For T5, we implement its fine-tuning on the Huggingface  transformers library on A100 GPUs. We use \texttt{T5}$_\text{LARGE}$~\citep{raffel2020t5} as our backbone.

\subsection{Experimental Details on DROP}
\noindent\textbf{Fine-tuning Details}~For all fine-tuning methods, we select the default max token length for each model. We set the max token length of  generation as $96$. To save training time, we set early stop mechanism: we evaluate the EM and F1 score per $500$ or $1000$ steps, if the performance does not increase in the latest $20$ evaluations, we stop the training and save the best checkpoint.

On DROP, we pre-pend the question to the given paragraph. For multi-span answer, we insert ``;'' between each span and make up the final answer. For \texttt{T5}$_\text{LARGE}$, we also insert ``$<$\texttt{/s}$>$'' token between the question and the given paragraph. Since most LMs' checkpoints on DROP is currently not off-the-shelf, we re-implement them and compare to the results reported in previous works. We present the comparison results in Table~\ref{tab:drop-reimplementation}.
\vspace{20pt}
\begin{table}[h]
\centering
\caption{Performance Comparison on DROP between reported results in previous works and our re-implementation. Results marked with $^*$ represent our re-implementation results.}
\label{tab:drop-reimplementation}
\scalebox{1.0}{%
    \begin{tabular}{lll}
    \toprule
    \small
    \textbf{Models} & \textbf{EM (\%)} & \textbf{F1 (\%)}  \\ \midrule
    BART~\citep{pi2022reasoning}    & $66.2$ & $69.2$\\
    BART$^*$                        &   $67.4$   &   $70.6$ \\
    \midrule
    T5~\citep{yoran2022turning} & --  & $64.6$\\
    T5$^*$ &   $61.0$        &   $64.6$ \\
    \midrule
    \poet-SQL~\citep{pi2022reasoning} & $77.7$ & $80.6$\\
    \poet-SQL$^*$ &   $76.9$   & $80.0$ \\
    \bottomrule
    \end{tabular}%
}
\end{table}
\vspace{20pt}

\section{More Results on DROP}\label{sec:full-results}

We present the performance breakdown of F1 on dev set of DROP in Table~\ref{tab:drop-by-type} Apart from fine-tuning models on DROP dataset, we also use GPT-3 to conduct a study on few-shot learning. We pre-pend 10 random training samples in train set, and run all cases where answer type equals to ``number''. We also apply our search-based algorithm on GPT-3. To save API calling time, we only substitute the number for one time. Table~\ref{tab:codex-few-shot-drop} presents the F1 score comparison.

We also summarize common calculation error cases in our tested language models and present some of them for case study in Table~\ref{tab:drop-error-case-study}, which again illustrates the unreliability of language models. 

\vspace{20pt}
\begin{table}[ht]
\centering
\small
\caption{Breakdown of model F1 score by answer types on the dev set of DROP.}
\label{tab:drop-by-type} 
\scalebox{1.0}{
\begin{tabular}{lccccc}
\toprule
\textbf{Models} & \textbf{Number} & \textbf{Span} & \textbf{Spans} & \textbf{Date} & \textbf{Total}  \\
\toprule
BART & $66.3$ &  $80.3$ & $66.0$ & $56.7$ & $70.6$\\
~~~~\textit{w.} \ours & $75.2$ & $80.5$ & $66.7$ & $55.7$ & $76.1$ \\ \hdashline
T5 & $55.5$ & $81.6$ & $73.0$ & $53.5$ & $64.6$ \\
~~~~\textit{w.} \ours & $69.8$ & $81.8$ & $73.9$ & $53.5$ & $73.5$\\ \hdashline
\tapex & $77.8$ & $84.3$ & $72.9$ & $62.8$ & $79.3$\\
~~~~\textit{w.} \ours & $81.4$ & $84.4$ & $73.0$ & $61.7$ & $81.6$ \\ \hdashline
\poet-SQL & $78.4$ &$84.6$ & $76.6$ & $63.4$ & $80.0$\\
~~~~\textit{w.} \ours & $81.4$ & $84.9$ & $76.9$ & $62.6$ & $82.0$ \\  
\bottomrule
\end{tabular}
}
\end{table}
\vspace{20pt}
\begin{table}[h]
    \small
    \centering
    \caption{Performance of GPT-3 \textit{w.} \ours on the DROP numeric subset.}
    \begin{tabular}{ccll}
        \toprule
        \textbf{Language Model} & \textbf{Algorithm} & {\textbf{F1(\%) on Hard}} & {\textbf{F1(\%) on Total}} \\
        \midrule
        \multirow{2}{*}{GPT-3~(175B)}          &  -  & $42.5$  & $64.7$  \\ 
                                &  Search  & $59.9$ {\footnotesize $(+17.4)$}  &   $68.7$  {\footnotesize $(+4.9)$}   \\
        \bottomrule
    \end{tabular}
    \label{tab:codex-few-shot-drop}
\end{table}
\vspace{30pt}
\begin{table}[h]
    \centering
    \caption{Common calculation error cases on DROP dataset.}
    \label{tab:drop-error-case-study}
    \scalebox{0.85}{
    \begin{tabular}{p{3.0cm}p{8.0cm}p{1.8cm}p{1.8cm}}
        \toprule    
        \multicolumn{1}{l}{\bf Error Type} & {\bf Example} & {\bf Prediction} & {\bf Label}\\ \midrule
        \multirow{1}{*}{Carry Error} & 
        \ldots the size of the black-white IQ gap in the United States decreased from {\color{Plum}\bf 16.33} to {\color{Plum}\bf 9.94} IQ points. \ldots &
        \multirow{1}{*}{6.49} & \multirow{1}{*}{6.39} \\
        & {\color{Bittersweet} \bf Q: }How many IQ points did the black-white IQ gap decrease in the United States in a 2013 analysis of the National Assessment of Educational Progress? & & \\ \midrule
        Missing High Digit & 
        \ldots The Department of Tourism recorded {\color{Plum}\bf26,861,095} Thai and {\color{Plum}\bf11,361,808} foreign visitors to Bangkok in 2010. \ldots &
        499287 & 15499287\\
        & {\color{Bittersweet} \bf Q: }How many more Thai visitors did Bangkok have in 2010 compared to other foreign visitors? & & \\ \midrule
        Extra Integer digit &
        \ldots Rayner nailed a {\color{Plum}\bf23}-yard field goal \ldots Rayner got a {\color{Plum}\bf54}-yarder and a {\color{Plum}\bf46}-yarder to end the half \ldots &
        111113 & 123 \\
        & {\color{Bittersweet} \bf Q:} How many total yards of field goals did Dave Rayner have? & & \\ \midrule
        Extra Float Number Digits &
       \ldots have estimated the IQ means of 17-year-old black, white, and Hispanic students to range respectively from {\color{Plum}\bf 90.45-94.15} \ldots &
        3.75 & 3.7 \\
        & {\color{Bittersweet} \bf Q: }How many points difference is the IQ range in 17-year-old black students? & & \\ \midrule
        Insufficient Precision & 
        \ldots The Diocese of Karelia has {\color{Plum}\bf 22,000} church members in {\color{Plum}\bf 12} parishes. \ldots&
        1833 & 1833.33 \\
        & {\color{Bittersweet} \bf Q:} How many church members approximately are in each one of the 12 parishes? & & \\
        \bottomrule
    \end{tabular}
    }
\end{table}
\vspace{20pt}
\end{document}